%% file: main.tex
\documentclass[letterpaper, 10 pt, conference]{ieeeconf}  
\usepackage{subfig}

\IEEEoverridecommandlockouts 

\input{commands}

\title{\LARGE \bf
\llamppi{}: Rapidly Adaptive Whole-body Control of Legged Robots with GPU-Accelerated Parallel Simulations}

\author{Sebin Jung$^{\bigstar}$%
\quad\quad\quad Maitham F. AL-Sunni$^{\bigstar,\ddag}$%
\quad\quad\quad Juan Alvarez-Padilla%
\\%
Zachary Manchester%
\quad\quad\quad Changliu Liu%
\quad\quad\quad John M. Dolan%
\thanks{S. Jung$^{\bigstar}$, M. F. AL-Sunni$^{\bigstar,\ddag}$, C. Liu, and J. M. Dolan are with Carnegie Mellon University, Pittsburgh, PA, USA. Email: \texttt{\{sebinj, malsunni, cliu6, jdolan\}@andrew.cmu.edu}}%
\thanks{J. Alvarez-Padilla and Z. Manchester are with Massachusetts Institute of Technology, Cambridge, MA, USA. Email: \texttt{\{jralvare, zacm\}@mit.edu}}%
\\[0.3em]%
{\it$^\bigstar$Co-first authors (equal contribution)\quad\quad\quad$^\ddag$Project lead}
}

\newcommand{\paperhero}{%
  \noindent
  \begin{minipage}{\linewidth}
    \centering
    \vspace{-8mm}
    \includegraphics[width=\linewidth]{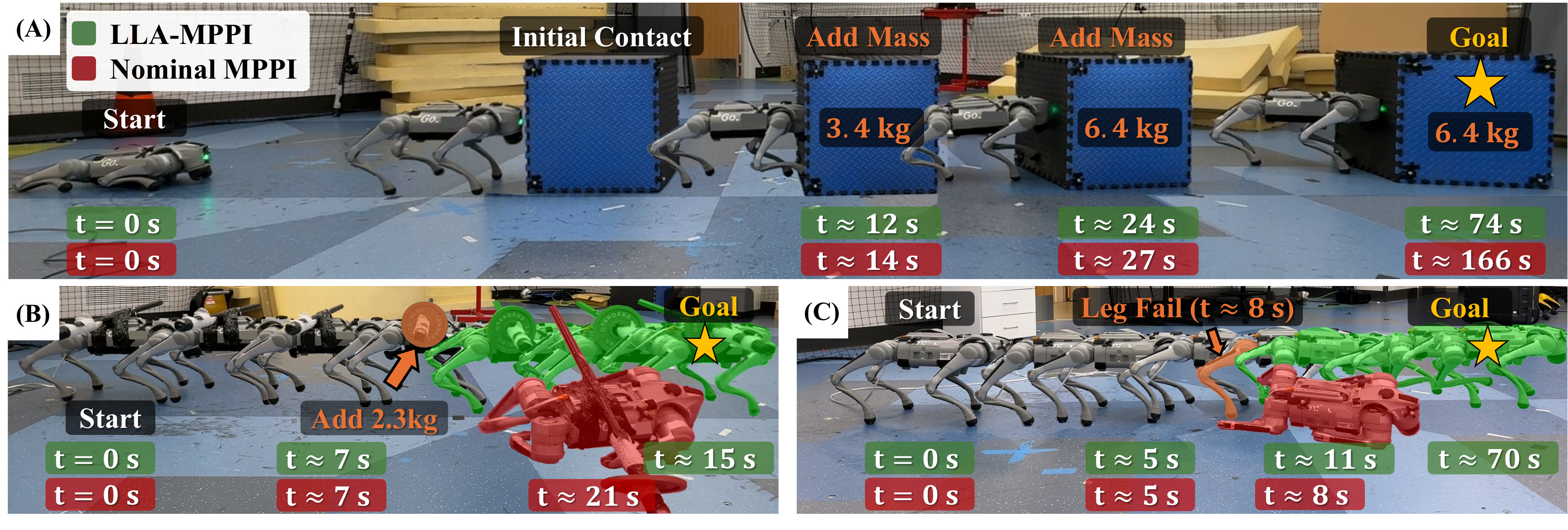}
    \vspace{-7mm}
    \captionof{figure}{%
      Hardware experiments on the Unitree Go2 with
\llamppi{} (\textcolor{llamppi}{green}) and nominal MPPI
(\textcolor{red}{red}). {\bf (A)}~The robot delivers a box to a goal
while mass is added on top of it on the fly; \llamppi{} finishes in
around $74\,\mathrm{s}$ versus $166\,\mathrm{s}$. {\bf (B)}~A
$2.3\,\mathrm{kg}$ payload is added on one side mid-walk; only
\llamppi{} reaches the goal. {\bf (C)}~The rear-left leg is disabled
mid-walk; only \llamppi{} reaches the goal. Videos of the full runs are on the
project's website.
    }
    \label{fig:headline}
    \vspace{0.5\baselineskip}
  \end{minipage}%
}

\IEEEaftertitletext{\paperhero}

\begin{document}

\maketitle

\thispagestyle{empty}
\pagestyle{empty}

\begin{abstract}
Real-time whole-body controllers for legged robots typically plan through a fixed nominal model and degrade when the deployed dynamics change. Adaptive methods typically require model structure that contact dynamics do not provide, or they need offline training for each anticipated condition. We present \textcolor{llamppi}{Look-back and Look-ahead Adaptive Model Predictive Path Integral control (LLA-MPPI)}. The method converts whole-body adaptation into selection over a bank of GPU-batched contact simulators with different physical or structural parameters. Windowed prediction errors select the simulator that best explains recent motion. A whole-body MPPI planner optimizes controls through the selected model. The framework requires no offline training, and its selected hypotheses are physically interpretable. Across four simulated tasks, it achieves $97.5\%$ success while the strongest baseline reaches $74\%$ and an oracle with the true model reaches $98.5\%$. Hardware validation on a Unitree Go2 shows the robot walking under a payload added mid-run, walking after one leg is disabled, and pushing a box to its goal while increasing its mass on the fly. Code, videos, and project details are available at:
\vspace{-5pt}
\begin{center}    
\href{https://lla-control.github.io}{\textcolor{llamppi}{https://lla-control.github.io}}
\end{center}
\end{abstract}

\vspace{-10pt}
\section{Introduction}
\label{sec:intro}

Legged robots encounter dynamics that change during operation:
payloads shift, ground friction varies, terrain tilts, and actuators
degrade. Real-time whole-body control has become practical
\cite{alvarez2024real,xue2025full,zhang2025whole,turrisi2024gpu},
but such controllers typically plan through a fixed nominal model and
degrade under model mismatch. Existing remedies either reduce the
model until a classical estimator applies, losing contact-mediated and
structural uncertainty, or train policies offline whose adaptation is
implicit and bounded by the training distribution.

\vspace{-1.7mm}
\subsection{Related Work}
\label{sec:related}

In adaptive control of legged systems, the identification model is
typically coarser than the control model. Estimators recover a small
number of inertial parameters on floating-base, single-rigid-body, or
centroidal models \cite{haack2025adaptive,sombolestan2024adaptive};
lump mismatch into an external wrench \cite{li2025whole}; or absorb it
into a black-box error model
\cite{zeng2024adaptive,zhou2025adaptive}. Within sampling-based
control, adaptation has instead targeted controller gains
\cite{11257687} or pre-designed gait modes \cite{11217651}, rather
than the dynamics model.

This reduction reflects the structure required by classical
estimators. Model-reference and $\mathcal{L}_1$ adaptive control,
recursive least squares, disturbance observers, and recursive Bayesian
filters respectively require matched uncertainty
\cite{narendra1989stable,hovakimyan2010l1}, an affine regressor
\cite{ljung1999system}, a discrepancy representable as a generalized
force \cite{deluca2003actuator}, or a fixed parameterization with
sufficient excitation \cite{gordon1993novel}. Whole-body contact
dynamics generally do not provide these structures, leaving
contact-mediated uncertainty such as foot--ground friction and
structural uncertainty such as actuator failure outside the
identification model.

Multiple-model adaptive control (MMAC) removes the regressor
requirement by scoring a finite set of candidate models against
observed data \cite{narendra1997adaptive}. Recent work recasts this
principle as massively parallel model selection for ground and aerial
vehicles. A large bank of models is scored using one-step prediction
error over a look-back window, and the selected model is passed to a
look-ahead predictive controller or safety filter. This approach was
introduced for autonomous racing in simulation
\cite{alsunni2025llampc}, subsequently deployed on an F1TENTH vehicle
\cite{liao2026llampc_embedded}, and extended to safety-critical
control of an F1TENTH vehicle and a Crazyflie quadrotor
\cite{alsunni2026lbacbf}. Unlike classical MMAC, these methods maintain
thousands of candidate models, score only recent transitions, and use
the selected model directly in the look-ahead stage rather than
choosing among pre-designed controllers.

However, extending this principle to whole-body legged control is not
straightforward. Prior formulations rely on low-dimensional models of
known functional form and on motion that sufficiently excites the
uncertain dynamics. Neither assumption holds generally for
contact-rich locomotion. The whole-body transition map changes across
contact modes and admits no single smooth closed-form description, while
candidate models may become indistinguishable whenever
motion does not excite the corresponding uncertainty.

A parallel line of work uses reinforcement learning instead of
explicit model-based adaptation. Domain randomization trains policies
across distributions of dynamics
\cite{lee2020learning,rudin2022learning}, teacher--student methods infer
latent environment representations \cite{kumar2021rma}, and mixtures
of specialized policies cover discrete failure modes
\cite{turrisi2026mixture}. These methods can perform well within their
training distributions, but additional anticipated modes generally
require retraining, the inferred latent is not a physical quantity,
and out-of-distribution degradation may occur without an explicit
indication of model mismatch.

To remove the model-structure assumptions above, we perform adaptation
directly over simulation-fidelity contact dynamics using GPU-batched
model selection. Candidate hypotheses need only be steppable by the
physics engine; no differentiable, affine, or closed-form
parameterization is required. GPU-parallel physics makes this practical
by advancing thousands of independently parameterized worlds
concurrently. We introduce
{\bf \textcolor{llamppi}{Look-back and Look-ahead Adaptive MPPI
(LLA-MPPI)}}, which maintains a bank of full-contact simulators in
MuJoCo Warp \cite{mujoco_warp_2025} spanning physical and structural hypotheses. A windowed
prediction residual selects the model that best explains recent motion,
while a switching margin retains the incumbent when intermittent
excitation makes hypotheses difficult to distinguish. A whole-body
MPPI controller \cite{williams2017mppi} then plans directly through
the selected model. Relative to prior multiple-model formulations,
LLA-MPPI removes the requirement for a low-dimensional model of known
functional form and regularizes online selection under intermittent
contact excitation. The method requires no offline policy training,
and its selected hypotheses retain direct physical meaning.

\subsection{Contributions}
\label{sec:contributions}

Our contributions are threefold:
\begin{enumerate}
    \item A whole-body adaptive control framework that performs online
    identification over a bank of full-contact simulators under
    intermittent contact excitation, covering parametric and structural
    uncertainty in a common hypothesis space without requiring
    differentiable, affine, or closed-form dynamics.

    \item A decoupled heterogeneous architecture that runs GPU-batched
    identification concurrently with CPU whole-body planning, so
    increasing model-set resolution primarily consumes GPU throughput
    rather than increasing the planner's critical-path latency.

    \item An evaluation across four contact-rich locomotion and
    loco-manipulation tasks against model-based adaptive-control
    baselines, together with a comparison to domain-randomized RL.
\end{enumerate}

\section{Problem Formulation}
\label{sec:problem_formulation}
We consider real-time whole-body control of a robot whose dynamics may
change during operation. Let
$\mathbf{x}_k = [\mathbf{q}_k^\top, \mathbf{v}_k^\top]^\top \in
\mathcal{X}$ denote the state at time step $k$, where $\mathbf{q}_k$
and $\mathbf{v}_k$ are the generalized configuration and velocity. For
loco-manipulation tasks, $\mathbf{x}_k$ also includes the states of
manipulated objects. The input
$\mathbf{u}_k \in \mathcal{U}$ is the vector of desired joint positions
tracked by low-level PD servos, where
$\mathcal{U} = [\mathbf{u}_{\min}, \mathbf{u}_{\max}]$ is the set of
admissible commands. These bounds limit the commands and do not by
themselves bound the actuator torques. The state evolves
as
\begin{align}
    \mathbf{x}_{k+1}
    =
    f_k\left(\mathbf{x}_k, \mathbf{u}_k\right),
    \label{eq:uncertain_dynamics}
\end{align}
and two properties of the transition map $f_k$ shape the problem.
First, $f_k$ describes whole-body motion under unilateral contact. The
map changes character whenever the active contact set changes (e.g., as
feet and manipulated objects make and break contact, and as frictional
contacts transition between sticking and slipping), so $f_k$ is
non-smooth in $(\mathbf{x}_k, \mathbf{u}_k)$ and admits no single
closed-form expression valid across contact modes. Second, $f_k$ is
unknown to the controller and piecewise constant in time. It may
change abruptly at unknown moments during an episode.

What is available at design time is an \emph{adaptation class}. A
parameterized family of candidate models
$\widetilde{f}(\mathbf{x}, \mathbf{u};\, \bm{\theta})$,
$\bm{\theta} \in \Theta$, chosen to cover the anticipated variation
and its physically plausible range. Elements of $\Theta$ may be
continuous physical parameters (e.g., payload mass, ground friction), or discrete structural conditions, such as leg
impairment. Each candidate inherits the contact structure of
\eqref{eq:uncertain_dynamics}. $\widetilde{f}$ is itself a non-smooth
transition map, required only to be evaluable forward in time, but not to
be differentiable, nor affine in $\bm{\theta}$. The true dynamics need
not coincide with any single element of the class, and the
controller's task is to determine which hypothesis currently explains
the robot's motion best, online and from measured transitions alone.

At each control step, the controller commits to a hypothesis
$\widehat{\bm{\theta}}_k \in \Theta$ and applies the first input of
the planned sequence
\begin{align}
\label{eq:adaptive_problem}
    \mathbf{U}_k^\star
    \in
    \operatorname*{arg\,min}_{\mathbf{U}_k}\
    & \sum_{h=0}^{H-1}
      \ell\left(\bar{\mathbf{x}}_{h+1}, \mathbf{u}_{h}, h\right), \\
    \mathrm{s.t.}\quad
    & \bar{\mathbf{x}}_{h+1}
      =
      \widetilde{f}\left(\bar{\mathbf{x}}_{h}, \mathbf{u}_{h};\,
      \widehat{\bm{\theta}}_k\right),
      \qquad
      \bar{\mathbf{x}}_{0} = \mathbf{x}_k, \nonumber \\
    & \mathbf{u}_{h} \in \mathcal{U},
      \qquad h = 0, \dots, H-1, \nonumber
\end{align}
where $\mathbf{U}_k = \{\mathbf{u}_{0}, \dots, \mathbf{u}_{H-1}\}$ is
a control sequence over the planning horizon $H$,
$\bar{\mathbf{x}}_{h}$ denotes states predicted by the planner, and
$\ell$ is the task stage cost, which may depend on the horizon step
$h$ through time-varying references. The input constraint reflects
actuation limits and is enforced exactly. Requirements on the motion
itself (e.g., tracking accuracy, posture regulation, task progress) are
encoded in $\ell$.

The problem addressed in this work is to choose
$\widehat{\bm{\theta}}_k$ rapidly enough to track both
parametric and structural changes, without policy training or offline
data collection, and at real-time rates---while
\eqref{eq:adaptive_problem} itself must be solved through non-smooth
contact dynamics that provide neither gradients nor convex structure
to exploit.

\section{Look-back and Look-ahead Adaptive MPPI}
\label{sec:llamppi}

\begin{figure*}[t]
    \centering
    \includegraphics[width=\linewidth]{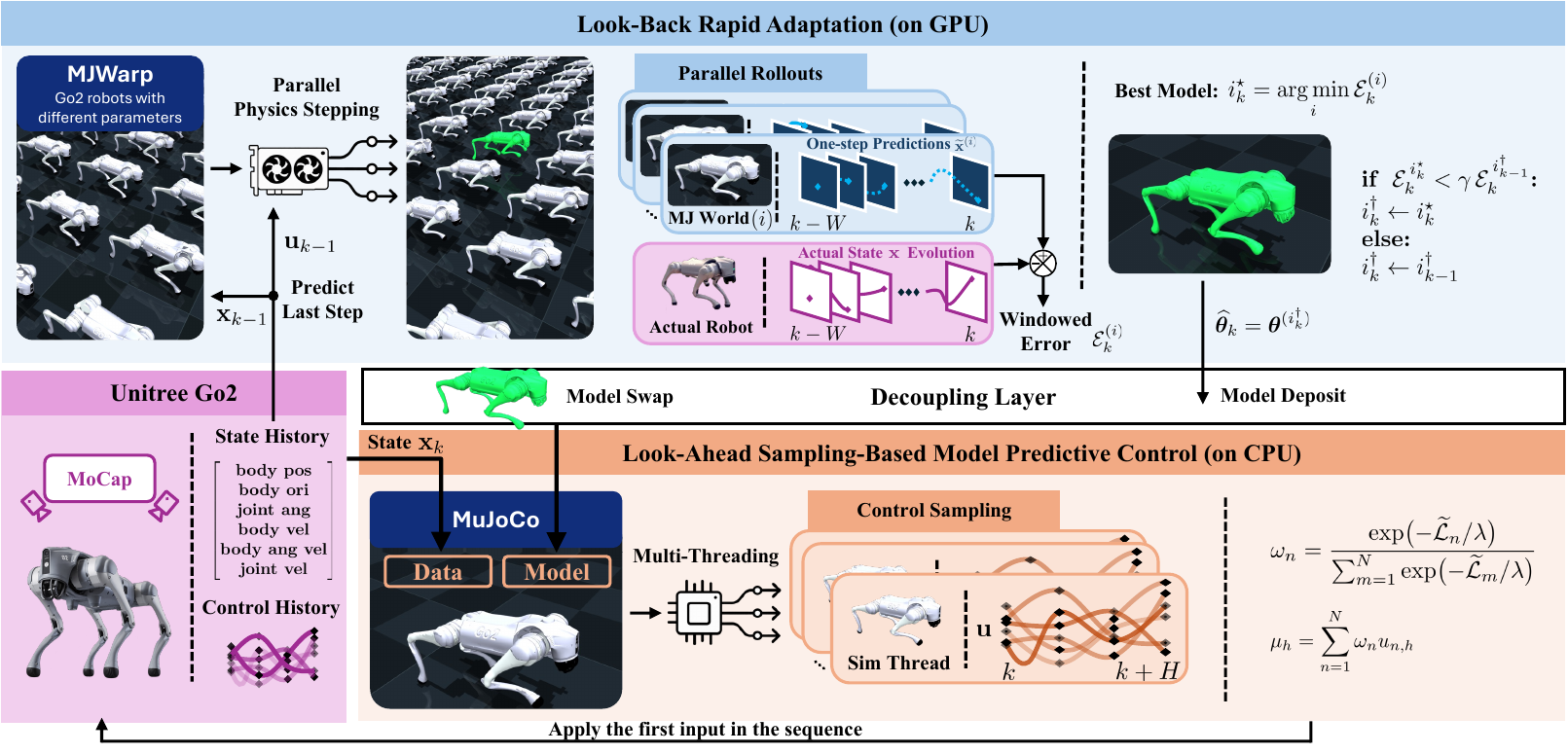}
    \caption{Overview of the \llamppi{} framework. The GPU-parallel look-back stage selects the model that best explains recent state transitions, while the CPU-parallel look-ahead stage uses the selected model to solve the MPPI. The two stages operate asynchronously to provide real-time adaptive control.}
    \label{fig:pipeline}
    \vspace{-13pt}
\end{figure*}

\begin{figure}[t]
    \centering
    \includegraphics[width=\columnwidth]{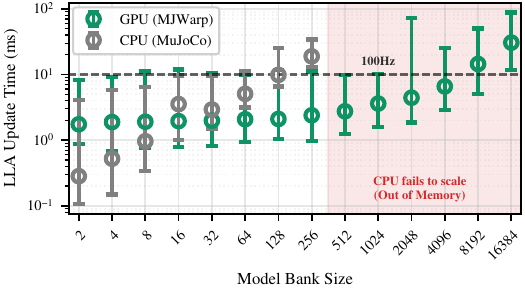}
    \caption{Look-back update time versus bank size $K$ for the
    GPU-batched MJWarp bank and a CPU MuJoCo baseline. Markers and
    error bars denote mean and spread across runs. Performed
    on an NVIDIA {RTX 4070 Laptop GPU} (8\,GB) and an AMD {Ryzen~9
    8945HS CPU} (8 cores / 16 threads).}
    \label{fig:lla_scale}
    \vspace{-13pt}
\end{figure}

\begin{algorithm}[t]
\footnotesize
\KwIn{bank $\Theta_K=\{\bm{\theta}^{(i)}\}_{i=1}^{K}$;
  index $i_{\mathrm{nom}}$ of the nominal hypothesis;
  residual weighting $S_{k}$; window $W$; switching margin $\gamma$;
  MPPI parameters $N$, $H$, $J$, $\lambda$, $\bm{\Sigma}_{\epsilon}$,
  $\rho$}
instantiate the bank $\mathcal{F}$ \eqref{eq:bank} as $K$ MJWarp worlds
  on the GPU\;
initialize the score buffers $\mathcal{E}^{(i)}_{0}\gets0,\ \forall i$\;
$i^{\dagger}_{-1}\gets i_{\mathrm{nom}}$,\ $\widehat{\bm{\theta}}_{-1}\gets\bm{\theta}^{(i_{\mathrm{nom}})}$\;
initialize the nominal sequence $\{\bm{\mu}_{h}\}_{h=0}^{H-1}$\;
\For{$k=0,1,2,\dots$}{
  measure $\mathbf{x}_{k}$\;
  \tcp{Look-back worker (GPU, batched)}
  \If{$k\geq1$}{
    broadcast $(\mathbf{x}_{k-1},\mathbf{u}_{k-1})$ to all $K$ worlds\;
    $\{\widetilde{\mathbf{x}}^{(i)}_{k}\}_{i=1}^{K}\gets
      \mathcal{F}_{\mathbf{x}_{k-1},\mathbf{u}_{k-1}}$
      \eqref{eq:bank_predictions}\;
    $r^{(i)}_{k}\gets$ weighted residual score,\ $\forall i$
      \eqref{eq:residual}\;
    $\mathcal{E}^{(i)}_{k}\gets\mathcal{E}^{(i)}_{k-1}
      +\bigl(r^{(i)}_{k}-r^{(i)}_{k-W}\bigr)/W,\ \forall i$
      \eqref{eq:window_argmin}\;
  }
  \eIf{$k\geq W$}{
    $i^{\star}_{k}\gets\operatorname*{arg\,min}_{i}
      \mathcal{E}^{(i)}_{k}$ \eqref{eq:window_argmin}\;
    \eIf{$\mathcal{E}^{(i^{\star}_{k})}_{k}<\gamma\,
      \mathcal{E}^{(i^{\dagger}_{k-1})}_{k}$ \eqref{eq:hysteresis}}{
      $i^{\dagger}_{k}\gets i^{\star}_{k}$\;
      synchronize $\widehat{\bm{\theta}}_{k}\gets
        \bm{\theta}^{(i^{\dagger}_{k})}$ with the planner\;
    }{
      $i^{\dagger}_{k}\gets i^{\dagger}_{k-1}$,\
        $\widehat{\bm{\theta}}_{k}\gets\widehat{\bm{\theta}}_{k-1}$\;
    }
  }{
    $i^{\dagger}_{k}\gets i^{\dagger}_{k-1}$,\
      $\widehat{\bm{\theta}}_{k}\gets\widehat{\bm{\theta}}_{k-1}$\;
  }
  \tcp{Look-ahead worker (CPU threads)}
  sample $\{\mathbf{u}^{(n)}_{h}\}_{n=1}^{N}$ by spline perturbation of
    $\{\bm{\mu}_{h}\}$ and clip to $\mathcal{U}$\;
  $\bar{\mathbf{x}}^{(n)}_{h+1}\gets
    \widetilde{f}(\bar{\mathbf{x}}^{(n)}_{h},\mathbf{u}^{(n)}_{h};\,
    \widehat{\bm{\theta}}_{k}),\
    \bar{\mathbf{x}}^{(n)}_{0}\gets\mathbf{x}_{k}$
    \eqref{eq:lookahead_rollouts}\;
  $\mathcal{L}_{n}\gets$ rollout cost,\ $\forall n$
    \eqref{eq:mppi_rollout_cost}\;
  $\omega_{n}\gets$ normalized weights,\ $\forall n$
    \eqref{eq:mppi_weights}\;
  $\bm{\mu}_{h}\gets\sum_{n=1}^{N}\omega_{n}\mathbf{u}^{(n)}_{h},\
    h=0,\dots,H-1$ \eqref{eq:mppi_update}\;
  apply $\mathbf{u}_{k}\gets\bm{\mu}_{0}$ and shift
    $\{\bm{\mu}_{h}\}$ forward\;
}
\caption{\llamppi{}}
\label{alg:lla_mppi}
\end{algorithm}

\llamppi{} solves the problem in Section \ref{sec:problem_formulation} using a single computational primitive: \emph{batched forward simulation} of contact dynamics. The framework maintains a bank of $K$ candidate physics models specified at design time, corresponding to $K$ hypotheses in $\Theta$. At every control step, it uses this primitive in two parallelized simulation workloads, as illustrated in \autoref{fig:pipeline}. In the \emph{look-back} stage, all candidates predict the transition most recently executed by the robot. A sliding window of one-step prediction errors selects the candidate that best predicts the observed transitions, yielding the parameter estimate $\widehat{\bm{\theta}}_k$. In the \emph{look-ahead} stage, an MPPI controller approximates \eqref{eq:adaptive_problem} by rolling out sampled control sequences through the selected model. In this way, online model estimation reduces to selection over the simulator bank: the candidate that best explains the recently observed transitions supplies the model parameterization used for subsequent planning rollouts.

\subsection{A Bank of Physics Models}
\label{sec:bank}

The candidate family of Section \ref{sec:problem_formulation} cannot be built from analytic models with the parameter structure that conventional estimators require. $\widetilde{f}$ must reproduce non-smooth whole-body contact dynamics, whose contact forces are themselves the output of a numerical solver. We therefore represent each candidate directly as a physics simulator.

Let $\mathcal{M}(\bm{\theta})$ denote a simulator instantiated with hypothesis $\bm{\theta}\in\Theta$, and let $\widetilde{f}(\mathbf{x},\mathbf{u};\,\bm{\theta})$ denote one integration step under $\mathcal{M}(\bm{\theta})$. We construct a finite hypothesis set $\Theta_K=\{\bm{\theta}^{(i)}\}_{i=1}^{K}$ by uniformly discretizing continuous parameter ranges, enumerating discrete structural conditions, and taking their Cartesian product. Instantiating one simulator per hypothesis yields the bank
\begin{align}
    \mathcal{F}
    :=
    \left\{
    \widetilde{f}(\cdot,\cdot\,;\bm{\theta}^{(i)})
    \right\}_{i=1}^{K}.
    \label{eq:bank}
\end{align}

All candidates share the same simulator structure and execution graph, while their physical or structural parameters vary across worlds. MJWarp represents the bank as $K$ worlds within one batched model, with per-world fields such as mass, inertia, friction, and actuator limits assigned at initialization. Dependent quantities are updated consistently with each hypothesis, such as scaling payload inertia with mass under fixed geometry. The complete simulation step is captured as a CUDA graph and replayed once per control step. Consequently, advancing the entire bank only requires one CUDA graph launch per control step, independent of $K$.

\subsection{Look-Back: Selection under Intermittent Excitation}
\label{sec:lookback}

As each new measured transition $(\mathbf{x}_k,\mathbf{u}_k,\mathbf{x}_{k+1})$ becomes available, the look-back stage produces one-step predictions in parallel by evaluating all $K$ candidate models from the measured pre-transition state $\mathbf{x}_k$:
\begin{align}
    \mathcal{F}_{\mathbf{x}_k,\mathbf{u}_k}
    =
    \left\{
        \widetilde{f}\!\left(
            \mathbf{x}_k,\mathbf{u}_k;\bm{\theta}^{(i)}
        \right)
    \right\}_{i=1}^{K}
    =
    \left\{
        \widetilde{\mathbf{x}}_{k+1}^{(i)}
    \right\}_{i=1}^{K}.
    \label{eq:bank_predictions}
\end{align}

Here $\mathcal{F}_{\mathbf{x},\mathbf{u}}$ denotes the bank evaluated at $(\mathbf{x},\mathbf{u})$. The measured state and applied command are broadcast to all candidate worlds, which are advanced by one batched MJWarp step. Each candidate's prediction error and scalar score are
\begin{align}
    \mathbf{e}_{k+1}^{(i)}
    =
    \widetilde{\mathbf{x}}_{k+1}^{(i)}-\mathbf{x}_{k+1},
    \qquad
    r_{k+1}^{(i)}
    =
    \sqrt{
        \bigl(\mathbf{e}_{k+1}^{(i)}\bigr)^\top
        S_{k+1}
        \mathbf{e}_{k+1}^{(i)}
    },
    \label{eq:residual}
\end{align}
where $S_{k+1}\succeq0$ is diagonal and specifies the state channels used for identification and their relative scaling. It may assign uniform weights, restrict the score to selected coordinates, or normalize heterogeneous channels using exponentially smoothed cross-candidate RMS errors, in which case it varies slowly over time.

Candidate predictions and residual reduction remain on the GPU, and only the $K$ scalar scores return to the CPU. To reduce sensitivity to individual contact events, the scores are averaged over a sliding window of $W$ transitions:
\begin{align}
    \mathcal{E}_k^{(i)}
    =
    \frac{1}{W}
    \sum_{\tau=k-W+1}^{k} r_{\tau}^{(i)},
    \qquad
    i_k^\star
    =
    \operatorname*{arg\,min}_{i}\mathcal{E}_k^{(i)}.
    \label{eq:window_argmin}
\end{align}
The scores are stored in a fixed-length buffer with $r^{(i)}_{\tau}=0$ for $\tau\leq0$, and selection begins once the window is full.

Let $i_{k-1}^{\dagger}$ denote the incumbent hypothesis. A challenger is accepted only if
\begin{align}
    \mathcal{E}_k^{(i_k^\star)}
    <
    \gamma\,
    \mathcal{E}_k^{(i_{k-1}^{\dagger})},
    \qquad
    \gamma\in(0,1],
    \label{eq:hysteresis}
\end{align}
and, when enabled, the same challenger satisfies this condition for a prescribed number of consecutive evaluations. Upon acceptance, $i_k^\dagger=i_k^\star$; otherwise, $i_k^\dagger=i_{k-1}^\dagger$. With $\gamma=1$ and persistence disabled, the rule reduces to immediate minimum-error selection while retaining the incumbent under exact ties. The active parameter estimate is therefore
$\widehat{\bm{\theta}}_k=\bm{\theta}^{(i_k^\dagger)}$. Before the window is full, for $k<W$, the incumbent is the nominal hypothesis $i_{\mathrm{nom}}$.

Candidate distinguishability depends on whether the executed motion excites the corresponding uncertainty. Friction becomes informative when contact mobilizes tangential forces, whereas an object's mass is unobservable while it remains stationary and out of contact. The look-back window and switching margin retain the incumbent during weakly informative phases, suppressing noise-driven switches while preserving a physically consistent planning hypothesis.

\subsection{Look-Ahead: Model Predictive Path Integral Control}
\label{sec:lookahead}

At control step $k$, let $\widehat{\bm{\theta}}_k$ denote the latest selected hypothesis available to the whole-body MPPI planner \cite{alvarez2024real}. Starting from the measured state $\bar{\mathbf{x}}_0^{(n)}=\mathbf{x}_k$, the planner propagates $N$ sampled control sequences over a horizon of $H$ steps:
\begin{align}
    \bar{\mathbf{x}}_{h+1}^{(n)}
    =
    \widetilde{f}\!\left(
        \bar{\mathbf{x}}_h^{(n)},
        \mathbf{u}_h^{(n)};
        \widehat{\bm{\theta}}_k
    \right),
    \quad
    \substack{h=0,\ldots,H-1,\\ n=1,\ldots,N.}
    \label{eq:lookahead_rollouts}
\end{align}

The planner maintains a nominal control sequence
$\{\bm{\mu}_h\}_{h=0}^{H-1}$. Following \cite{alvarez2024real,howell2022predictive},
temporally smooth samples are generated by perturbing $J$ control
points of the nominal sequence with zero-mean Gaussian noise of
diagonal covariance $\bm{\Sigma}_{\epsilon}$ and evaluating the
resulting cubic spline along the horizon. The sampled
commands are then clipped to the admissible set $\mathcal{U}$.

Each rollout is evaluated using the task-dependent stage cost $\ell(\mathbf{x},\mathbf{u},h)
    =
    \ell_{\mathrm{track}}\!\left(
        \mathbf{x},\mathbf{x}_{\mathrm{ref},h}
    \right) +
    \ell_{\mathrm{cmd}}(\mathbf{x},\mathbf{u})
    +
    \phi_{\mathrm{task}}(\mathbf{x},h)$.
Here,
$\ell_{\mathrm{track}}
=
\|\mathbf{e}_{\mathrm{pose}}\|_{Q}^{2}
+
\|\mathbf{p}-\mathbf{p}_{\mathrm{ref}}\|_{1,Q_p}$
combines a quadratic penalty on $\mathbf{e}_{\mathrm{pose}}$, which stacks the quaternion, joint, and velocity tracking errors, with the $Q_p$-weighted $\ell_1$ error of the base position $\mathbf{p}$. The command cost is
$\ell_{\mathrm{cmd}}=\|\bm{\tau}_{\mathrm{PD}}\|_{R}^{2}$, where
$\bm{\tau}_{\mathrm{PD}}
=
K_p(\mathbf{u}-\mathbf{q}_{\mathrm{jnt}})
-
K_d\dot{\mathbf{q}}_{\mathrm{jnt}}$
penalizes the commanded PD torque, with measured joint states
$\mathbf{q}_{\mathrm{jnt}}$ and servo gains $K_p$ and $K_d$.
The term $\phi_{\mathrm{task}}$ contains task-specific objectives, such as the weighted $\ell_1$ object-position error used for loco-manipulation. The rollout cost is
\begin{multline}
    \mathcal{L}_n
    =
    \sum_{h=0}^{H-1}
    \ell\!\left(
        \bar{\mathbf{x}}_{h+1}^{(n)},
        \mathbf{u}_h^{(n)},h
    \right) \\
    {}+
    (\rho-1)
    \ell\!\left(
        \bar{\mathbf{x}}_{H}^{(n)},
        \mathbf{u}_{H-1}^{(n)},H-1
    \right),
    \label{eq:mppi_rollout_cost}
\end{multline}
where $\rho\geq1$ is a task-dependent terminal multiplier. Setting $\rho=1$ recovers the unweighted rollout cost, while $\rho>1$ emphasizes the final stage.

The rollout costs are min-max normalized as
$\widetilde{\mathcal{L}}_n
=
(\mathcal{L}_n-\mathcal{L}_{\min})
/(\mathcal{L}_{\max}-\mathcal{L}_{\min})$, with
$\mathcal{L}_{\min}=\min_{n}\mathcal{L}_{n}$ and
$\mathcal{L}_{\max}=\max_{n}\mathcal{L}_{n}$, and converted into
exponential weights
\begin{align}
    \omega_n
    =
    \frac{\exp\bigl(-\widetilde{\mathcal{L}}_n/\lambda\bigr)}
    {\sum_{m=1}^{N}\exp\bigl(-\widetilde{\mathcal{L}}_m/\lambda\bigr)},
    \label{eq:mppi_weights}
\end{align}
where $\lambda>0$ is the temperature and $\omega_n=1/N$ when
$\mathcal{L}_{\max}=\mathcal{L}_{\min}$. The nominal sequence is then
updated as
\begin{align}
    \bm{\mu}_h
    =
    \sum_{n=1}^{N}
    \omega_n\mathbf{u}_h^{(n)},
    \qquad
    h=0,\ldots,H-1.
    \label{eq:mppi_update}
\end{align}
The first command $\bm{\mu}_0$ is applied, and the updated sequence is shifted forward to warm-start the next planning step. Because $\mathcal{U}$ is a box and the samples are clipped into it, this weighted mean remains admissible, so the applied command satisfies the input constraint of \eqref{eq:adaptive_problem} exactly.

The selected hypothesis is transferred to the planner without fitting, linearization, or projection onto a surrogate dynamics class. Continuous hypotheses directly update the physical fields of the MuJoCo rollout model, whereas discrete structural hypotheses replace the planner's compiled model and remain per-world field changes within the batched bank. When specified by a hypothesis, associated cost weights or sampling parameters are also synchronized, while the task reference and cost structure remain fixed.

\subsection{Computational Structure}
\label{sec:compute}

The \emph{look-back} and \emph{look-ahead} stages have complementary computational structures. The \emph{look-back} workload is wide and shallow, comprising $K$ independent one-step simulations initialized from the same measured state and control input and evaluated as parallel worlds in MJWarp \cite{mujoco_warp_2025}. In contrast, the \emph{look-ahead} workload is narrow and deep. $N$ trajectories are evaluated in parallel across CPU worker threads, while each trajectory advances sequentially through the $H$-step planning horizon under the selected model in MuJoCo \cite{todorov2012mujoco}. The architecture therefore assigns each workload to the computing resource suited to its parallel structure.

The two workloads are decoupled to keep identification outside the real-time planning path. The GPU model bank runs asynchronously in a background worker using the latest measured transition, while the CPU planner proceeds with the most recently completed model selection. The selection used by the planner may therefore lag the newest transition by up to one look-back cycle. Model changes are synchronized between rollout calls so that each MPPI update uses a consistent simulator. Increasing the bank resolution enlarges the parallel GPU batch without increasing its simulation depth. Consequently, it primarily consumes GPU throughput rather than MPPI solve time. As shown in \autoref{fig:lla_scale}, the bank evaluation time remains nearly constant with $K$ until the GPU becomes saturated.

\section{Experimental Validation}
\label{sec:experiments}

\begin{figure}[t]
    \centering
    \includegraphics[width=\columnwidth]{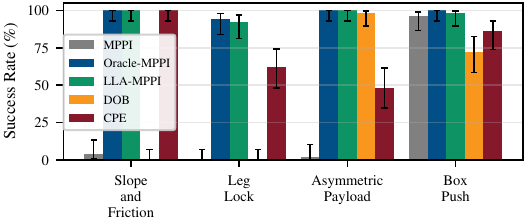}
    \caption{Success rates over 50 randomized trials per task. Error bars denote 95\% Wilson confidence intervals. Success is defined as reaching the task goal within the episode.}
    \label{fig:success_rate}
    \vspace{-20pt}
\end{figure}

\begin{figure*}[t]
    \centering
    \includegraphics[width=0.9\linewidth]{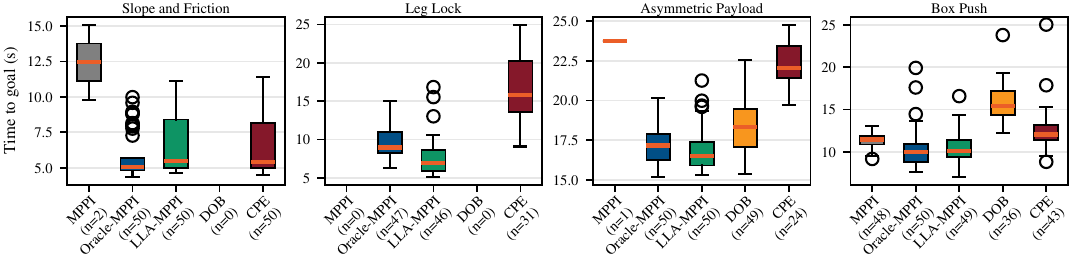}
    \caption{Time-to-goal on successful trials only (failed runs are excluded). Boxes show the interquartile range (IQR), red horizontal lines indicate medians, whiskers extend to the most extreme points within $1.5\times$ IQR. Sample size $n$ is the number of successful trials out of 50.}
    \label{fig:time_to_goal}
\end{figure*}

\begin{table*}[t]
\centering
\begin{tabular}{c|ccccc}
\toprule
\textbf{Method}
  & \textbf{Success rate} (\%)
  & \textbf{Tasks} $\geq 90\%$
  & \textbf{Fall rate} (\%)
  & \textbf{Solve time} (ms)
  & \textbf{Explicit ID} \\
\midrule
\mppi
  & $25.5$ [$20.0$, $32.0$]
  & $1/4$
  & $65.3$ [$57.4$, $72.5$]
  & $16.57$
  & --- \\
\oracle
  & \textcolor{gray}{$\mathbf{98.5}$ [$95.7$, $99.5$]}
  & \textcolor{gray}{$\mathbf{4/4}$}
  & \textcolor{gray}{$\mathbf{0.0}$ [$0.0$, $2.5$]}
  & \textcolor{gray}{$17.77$}
  & \textcolor{gray}{true model (given)} \\
\llamppi
  & $\mathbf{97.5}$ [$94.3$, $98.9$]
  & $\mathbf{4/4}$
  & $\mathbf{0.0}$ [$0.0$, $2.5$]
  & $17.60$
  & discrete bank \\
\dob
  & $42.5$ [$35.9$, $49.4$]
  & $1/4$
  & $33.3$ [$26.3$, $41.2$]
  & $\mathbf{15.43}$
  & --- \\
\cpe
  & $74.0$ [$67.5$, $79.6$]
  & $1/4$
  & $21.3$ [$15.5$, $28.6$]
  & $18.54$
  & continuous $\theta$ \\
\bottomrule
\end{tabular}
{\small
\caption{Aggregate results over four tasks with 50 trials each (200
episodes per method). All trials were run on a single workstation (Intel Core {i9-9980XE}, 18 cores / 36 threads; NVIDIA {RTX 2080 Ti}, 11\,GB; Ubuntu 22.04; MuJoCo 3.4.0 with MuJoCo-Warp 1.12.0), using one GPU per run. Brackets denote 95\% Wilson confidence intervals. Fall rates exclude the leg-lock task, where falls are not defined by the protocol. Solve time is the mean synchronous controller-update latency, and Explicit ID indicates whether a method produces a physical parameter estimate. Bold marks the best value per column among the realizable methods, and the gray row is the oracle reference.}
\label{tab:overall_comparison}
}
\vspace{-15pt}
\end{table*}

\begin{figure}[t]
    \centering
    \includegraphics[width=\columnwidth]{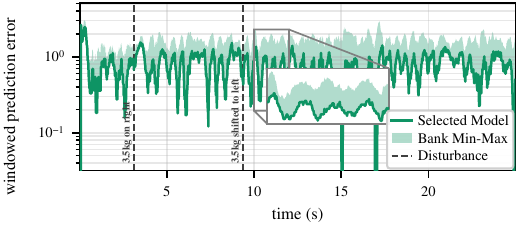}
    \caption{Windowed one-step prediction error during the \emph{asymmetric-payload} task. The shaded region spans the minimum-to-maximum error across the model bank, and dashed lines indicate payload changes.}
    \label{fig:lla_pred_error}
    \vspace{-15pt}
\end{figure}

\begin{figure}[h]
    \centering
    \includegraphics[width=\columnwidth]{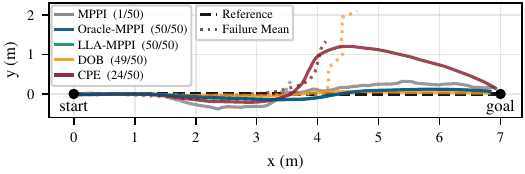}
    \includegraphics[width=\columnwidth]{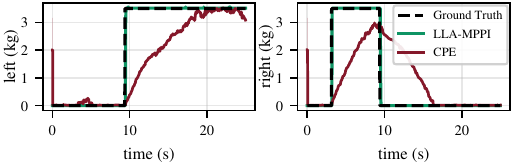}
    \caption{\emph{Asymmetric-payload} results. Top: mean planar trajectories relative to the reference path, with parentheses indicating successful trials out of 50 and dotted curves denoting mean failed trajectories. Bottom: estimated left and right payload masses from \llamppi{} and CPE.}
    \label{fig:id_payload}
    \vspace{-15pt}
\end{figure}

\begin{figure}[h]
    \centering
    \includegraphics[width=\columnwidth]{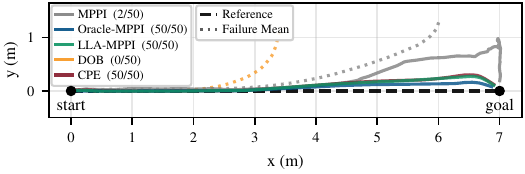}
    \includegraphics[width=\columnwidth]{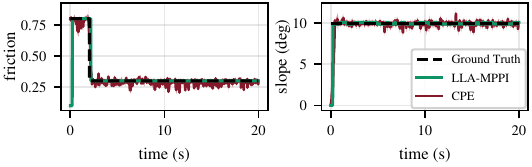}
    \caption{Results for the \emph{slope and friction} task. Top: mean planar trajectories. Bottom: friction and terrain-slope estimates relative to ground truth. \llamppi{} and CPE track both changes and complete all trials, whereas nominal MPPI and DOB largely fail.}
    \label{fig:id_friction_slope}
    \vspace{-15pt}
\end{figure}

In this section, we evaluate \llamppi{} through two complementary comparisons. First, we compare it with fixed-model MPPI, two online adaptation baselines, and an oracle across four simulated dynamics variations: \emph{asymmetric payload}, \emph{leg lock}, \emph{variable-mass box pushing}, and coupled \emph{slope and friction} variation. The oracle is always given the true model and therefore represents instant adaptation, the perfect-model reference that any adaptive method aims for. This study comprises $50$ randomized trials per task and controller, totaling $1{,}000$ simulated episodes. We further compare \llamppi{} with fixed-model MPPI on hardware for the payload, leg-lock, and box-pushing tasks. Second, we compare \llamppi{} with domain-randomized PPO to examine performance relative to offline-trained robustness.

All simulated experiments use the Unitree Go2 model in MuJoCo with the controller running online in real time. The ground-truth environment applies task-specific dynamics changes that are not disclosed to the controllers. Within each task, all MPPI-based methods are initialized with the same planning configuration, but subsequent model-dependent updates follow each method's adaptation mechanism. All realized conditions lie within the corresponding model-bank ranges, although continuous values are not required to coincide with discrete bank points.

\subsection{Comparison with Model-Based Adaptation Baselines}
\label{sec:adaptive_comparison}

We compare \llamppi{} with three baselines and an oracle. Nominal
MPPI retains its initial dynamics model throughout the episode.
\oracle{} implements the oracle as an MPPI planner that receives the
true model at every step, including each scheduled change.
\textcolor{dob}{Disturbance-observer-based MPPI (DOB)} estimates a
lumped disturbance from the whole-body inverse-dynamics
residual between the generalized force that the nominal model requires
to reproduce the measured motion and the applied actuator
force \cite{491410,10731682}. The residual is low-pass filtered and
its six floating-base components are saturated to form an estimated
external wrench. The wrench is held constant along each MPPI rollout.
\textcolor{dob}{DOB} therefore compensates additively without
changing the rollout model parameters.

The \textcolor{cpe}{continuous parameter estimator (CPE)} is a
bootstrap particle filter over the continuous parameter
space \cite{gordon1993novel}. Each particle parameterizes one world in
the same batched MJWarp simulator used by \llamppi{} and evolves
through a bounded Gaussian random walk. At every step the particles
are reweighted according to their one-step prediction residuals and
are resampled when the effective sample size falls below a threshold.
The posterior mean is then transferred to the MPPI rollout model.
\textcolor{cpe}{CPE} therefore uses the same number of simulator
worlds and the same residuals as \llamppi{} but maintains a continuous
estimate instead of selecting from a fixed hypothesis bank.

For the \emph{asymmetric payload}, \emph{leg lock}, and \emph{variable-mass box pushing} tasks, paired hardware demonstrations further illustrate \llamppi{} against \textcolor{mppi}{nominal MPPI} on the Unitree Go2 (\autoref{fig:headline}). On hardware, OptiTrack measures the base position, orientation, and linear and angular velocities, while onboard encoders provide the joint states. OptiTrack also tracks the box during pushing. Specific cost configurations are used, but no model parameters are calibrated offline.

\subsubsection{Asymmetric payload}
In simulation, the robot walks toward a goal while an initially balanced payload becomes asymmetric and subsequently shifts between sides of the body. The model bank varies the left and right payload masses to represent both the magnitude and spatial distribution of the inertial change. The bank comprises $K=10{,}000$ hypotheses on a $100\times100$ grid over the left- and right-side payload masses, each spanning $[0,4]\,\mathrm{kg}$. Model selection uses $W=20$, $\gamma=0.999$, and RMS-normalized configuration-and-velocity residuals, with no persistence requirement. \autoref{fig:lla_pred_error} shows that the selected hypothesis follows the model with the lowest prediction error, while \autoref{fig:id_payload} shows that \llamppi{} adapts faster than \textcolor{cpe}{CPE} after each transition. On hardware, \llamppi{} reaches the goal after a $2.3\,\mathrm{kg}$ right-side payload is added mid-run, whereas \textcolor{mppi}{nominal MPPI} falls toward the loaded side, as shown in \autoref{fig:headline}{\bf (B)}.

\subsubsection{Leg lock}
One leg becomes mechanically restricted during locomotion without an explicit fault signal. This disturbance changes the effective actuation structure and cannot be represented solely as an additive external wrench. The model bank therefore includes distinct hypotheses for the possible impaired legs, allowing recent prediction errors to determine the compatible structural model. The bank comprises $K=5$ hypotheses: the nominal model and one variant per leg with its three actuators disabled. Selection uses $W=10$, $\gamma=0.5$, three-step persistence, and unnormalized joint-position residuals. On hardware, \llamppi{} maintains balance and reaches the goal after the rear-left leg is disabled, whereas \textcolor{mppi}{nominal MPPI} falls immediately toward the impaired leg, as shown in \autoref{fig:headline}{\bf (C)}.

\subsubsection{Slope and friction}
The robot traverses a fixed incline while the foot-ground friction coefficient decreases during the episode. The model bank spans both terrain inclination and friction, requiring it to distinguish coupled changes in gravitational loading and tangential contact behavior. The bank comprises $K=10{,}000$ hypotheses on a $100\times100$ grid over terrain inclination in $[0,20]^{\circ}$ and friction coefficient in $[0.1,1.0]$. Selection uses $W=20$, $\gamma=0.99$, and RMS-normalized configuration-and-velocity residuals, with no persistence requirement. The identification traces in \autoref{fig:id_friction_slope} show that \llamppi{} selects hypotheses consistent with the active conditions and retains its incumbent during less informative phases of locomotion.

\subsubsection{Variable-mass box pushing}
The robot pushes a box whose mass increases over successive segments, extending adaptation to uncertain object dynamics coupled through intermittent contact. The bank comprises $K=100$ models with box mass uniformly spanning $[1,15]\,\mathrm{kg}$ and fixed geometry and friction. Selection uses $W=20$, $\gamma=0.999$, and unnormalized configuration residuals, with no persistence requirement. The rollout objective augments whole-body tracking with a weighted $\ell_1$ box-position error; for \llamppi{}, this weight is scaled by the ratio of the selected to nominal box mass. \llamppi{} responds rapidly to mass transitions, while CPE converges more gradually, consistent with the aggregate results in \autoref{tab:overall_comparison}. On hardware, \llamppi{} completes the task substantially faster than nominal MPPI under two successive increases in box mass, as shown in \autoref{fig:headline}{\bf (A)}.

The aggregate simulation results are reported in \autoref{fig:success_rate}, \autoref{fig:time_to_goal}, and \autoref{tab:overall_comparison}. Across the four tasks, \llamppi{} achieves a $97.5\%$ success rate, compared with $25.5\%$ for \textcolor{mppi}{nominal MPPI}, $42.5\%$ for \textcolor{dob}{DOB}, and $74.0\%$ for \textcolor{cpe}{CPE}. \oracle{} succeeds in $98.5\%$ of episodes, so \llamppi{} is within one percentage point of the oracle, with overlapping confidence intervals. Among the realizable methods, \llamppi{} is the only one to reach at least $90\%$ success on every task. Its mean controller update time is $17.60\,\mathrm{ms}$, only $1.03\,\mathrm{ms}$ above \textcolor{mppi}{nominal MPPI}, because the model-bank evaluation runs outside the controller's critical path.

\begin{figure}[t]
    \centering
    \includegraphics[width=\columnwidth]{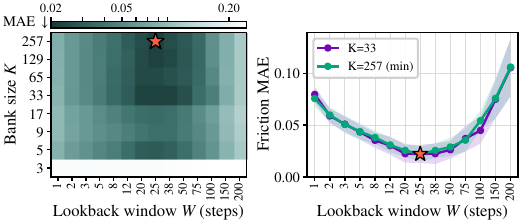}
    \caption{Ablation of lookback window $W$ and bank size $K$ on the \emph{slope-friction} task with noisy state measurements. Results show mean friction-estimation error over 10 paired seeds, with 95\% confidence intervals. The star marks the minimum over the tested settings at $(K,W)=(257,25)$.}
    \label{fig:ablation_window_bank}
    \vspace{-10pt}
\end{figure}

\autoref{fig:ablation_window_bank} ablates the look-back window $W$ and the bank size $K$ on the \emph{slope and friction} task with noisy state measurements. The friction-estimation error follows a U shape in $W$. A short window makes the identification sensitive to measurement error, while a long window lets outdated data from before a dynamics change influence the selection. An intermediate window balances the two. Refining the bank reduces the error further, with the best setting at $(K,W)=(257,25)$, close to the $W=20$ used in the main experiments.

\subsection{Comparison with Domain-Randomized RL}
\label{sec:ppo_comparison}

\begin{figure}[t]
    \centering
    \includegraphics[width=\columnwidth]{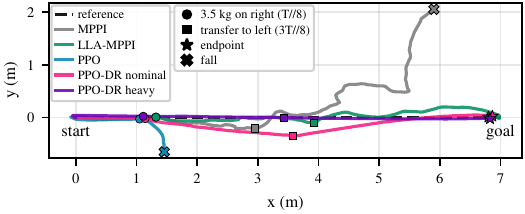}
    \caption{Representative trajectories for the \emph{asymmetric-payload} RL comparison. Markers show the $3.5\,\mathrm{kg}$ right-side payload addition, its transfer to the left, endpoints, and falls. \llamppi{} and both PPO-DR policies reach the goal, whereas nominal MPPI and PPO fall.}
    \label{fig:rl_payload}
    \vspace{-17pt}
\end{figure}

We compare \llamppi{} with three PPO policies on the \emph{asymmetric-payload} task: one trained without payload randomization (PPO), and two trained under payload domain randomization of different breadths---PPO-DR (nominal) and PPO-DR (heavy), which randomize the left and right payload masses independently over $[0,3]\,\mathrm{kg}$ and $[0,5]\,\mathrm{kg}$ per side, respectively. Nominal MPPI is included as a reference. All methods traverse the same waypoint sequence and are evaluated in the same simulator with the same task geometry, payload schedule, and success, completion-time, and fall criteria, while each method uses its own control interface. For PPO, the active waypoint is converted into the body-velocity and yaw-rate command used during training. This experiment contrasts explicit online model selection with robustness learned through offline domain randomization.

Representative trajectories are shown in \autoref{fig:rl_payload}. As the payload changes in mass and lateral placement, \llamppi{} updates its rollout model from observed transitions. PPO receives no payload estimate and must accommodate these changes through robustness encoded in its policy.

The computational requirements are summarized in \autoref{tab:offline_cost}. PPO requires no online trajectory optimization, but each policy is trained using $200$ million simulated steps. In contrast, \llamppi{} requires no offline policy training and performs model selection and trajectory optimization online. The table also reports the training samples required to reach each fraction of the final mean velocity-tracking return $\bar{R}_{\mathrm{vel}}^{\mathrm{final}}$. For each evaluation rollout, $R_{\mathrm{vel}}=\sum_{t=0}^{999}\exp\!\left(-\|\mathbf{v}^{\mathrm{cmd}}_{xy,t}-\mathbf{v}^{\mathrm{body}}_{xy,t}\|_2^2/0.25\right)$, where $\mathbf{v}^{\mathrm{cmd}}_{xy,t}$ and $\mathbf{v}^{\mathrm{body}}_{xy,t}$ are the commanded and measured planar body velocities; the bar denotes averaging across evaluation rollouts, and ``final'' denotes the final training checkpoint.

\begin{table}[h]
\centering
\footnotesize
\setlength{\tabcolsep}{3pt}
\begin{tabular}{c|ccccc}
\toprule
\textbf{Method} & \textbf{Offline}
  & \textbf{90\%} & \textbf{95\%} & \textbf{99\%}
  & \textbf{Online} \\
\midrule
PPO-DR (nominal)
  & $48.5$\,min & $46$M & $46$M & $69$M & $<1$\,ms \\
PPO-DR (heavy)
  & $54.7$\,min & $92$M & $115$M & $161$M & $<1$\,ms \\
\llamppi
  & -- & -- & -- & -- & $14.4 \pm 1.2$\,ms \\
\bottomrule
\end{tabular}
\caption{Training and online computational cost for the \emph{asymmetric payload} comparison. The percentage columns report the first PPO checkpoint reaching the indicated fraction of its final mean velocity-tracking return. Offline denotes PPO training wall time on an {RTX 4070 Laptop GPU}; online denotes computation per control step, including MPPI and model-bank computation for \llamppi{}.}
\label{tab:offline_cost}
\end{table}

\vspace{-10pt}
\section{Conclusions and Future Work}
\label{sec:conclusions}

We presented \llamppi{}, a whole-body adaptive control framework operating directly on simulation-fidelity contact dynamics. It combines GPU-batched model selection with a decoupled MPPI planner. Simulation experiments and hardware demonstrations showed adaptation to continuous physical variations and discrete structural changes without closed-form dynamics or offline policy training.

The method is limited by its finite, design-time hypothesis bank and by the distinguishability of candidate models from recent motion. Future work includes adaptive bank refinement, uncertainty-aware planning over multiple candidates, and hardware validation under diverse contact conditions.

\printbibliography
\end{document}

%% file: commands.tex
\usepackage{amsmath,amsfonts}
\usepackage{graphicx}
\usepackage{textcomp}
\usepackage{comment}

\usepackage{amsthm}
\usepackage[utf8]{inputenc}
\usepackage[T1]{fontenc}
\usepackage{enumerate}
\usepackage{soul}
\usepackage[dvipsnames,table]{xcolor} 
\usepackage{hyperref}
\usepackage{siunitx}
\usepackage{multirow}
\usepackage{subfig}
\usepackage{booktabs}
\usepackage[ruled,vlined]{algorithm2e}
\usepackage{algpseudocode}
\DontPrintSemicolon

\usepackage{amssymb}    
\usepackage{mathtools}  
\usepackage{bm}         
\usepackage{accents}
\usepackage{caption}
\usepackage{colortbl}  
\usepackage{array}     
\usepackage{adjustbox} 
\usepackage{wasysym}  
\usepackage{pifont}
\usepackage{stfloats}
\usepackage{url}
\usepackage{verbatim}
\usepackage{makecell}
\usepackage{tikz}
\usepackage{color}  
\usepackage{float}

\usepackage[style=ieee,backend=biber,sortcites=true]{biblatex}

\usepackage{fontawesome5}

\usepackage{cuted}

\renewcommand*{\multicitedelim}{\addcomma\space}
\DeclareCiteCommand{\cite}[\mkbibbrackets]
  {\usebibmacro{prenote}}
  {\usebibmacro{citeindex}\printfield{labelnumber}}
  {\multicitedelim}
  {\usebibmacro{postnote}}
  
\AtBeginBibliography{\scriptsize}  
\definecolor{blue}{HTML}{4E95D9}
\definecolor{gold}{HTML}{E5AE1C}
\definecolor{safe}{HTML}{4169E1}
\definecolor{unsafe}{HTML}{C41230}

\AtEveryBibitem{%
  \clearfield{doi}\clearfield{url}\clearfield{urlyear}\clearfield{urlmonth}\clearfield{urlday}%
  \clearfield{isbn}\clearfield{issn}\clearfield{eprint}\clearfield{month}%
  \clearlist{language}%
  \ifentrytype{inproceedings}{%
    \clearlist{publisher}\clearlist{organization}\clearlist{location}%
    \clearname{editor}\clearfield{series}\clearfield{volume}}{}%
}

\definecolor{llamppi}{HTML}{0E9464}  
\definecolor{mppi}{HTML}{000000}
\definecolor{dob}{HTML}{000000}
\definecolor{cpe}{HTML}{000000}
\definecolor{oracle}{HTML}{000000}
\newcommand{\llamppi}{{\bfseries\textcolor{llamppi}{LLA-MPPI}}}

\newcommand{\oracle}{\textcolor{oracle}{Oracle-MPPI}}
\newcommand{\mppi}{{\textcolor{mppi}{MPPI}}}
\newcommand{\dob}{{\textcolor{dob}{DOB}}}
\newcommand{\cpe}{{\textcolor{cpe}{CPE}}}

\hypersetup{hidelinks}